# Micro-Facial Expression Recognition Based on Deep-Rooted Learning Algorithm

**S. D. Lalitha**

Assistant Professor, R.M.K Engineering College, Chennai, India

**K. K. Thyagharajan**

Professor & Dean (Academic), Department of ECE, R.M.D Engineering College, Chennai, India

ABSTRACT

Facial expressions are important cues to observe human emotions. Facial expression recognition has attracted many researchers for years, but it is still a challenging topic since expression features vary greatly with the head poses, environments, and variations in the different persons involved. In this work, three major steps are involved to improve the performance of micro-facial expression recognition. First, an Adaptive Homomorphic Filtering is used for face detection and rotation rectification processes. Secondly, Micro-facial features were used to extract the appearance variations of a testing image-spatial analysis. The features of motion information are used for expression recognition in a sequence of facial images. An effective Micro-Facial Expression Based Deep-Rooted Learning (MFEDRL) classifier is proposed in this paper to better recognize spontaneous micro-expressions by learning parameters on the optimal features. This proposed method includes two loss functions such as cross entropy loss function and centre loss function. Then the performance of the algorithm will be evaluated using recognition rate and false measures. Simulation results show that the predictive performance of the proposed method outperforms that of the existing classifiers such as Convolutional Neural Network (CNN), Deep Neural Network (DNN), Artificial Neural Network (ANN), Support Vector Machine (SVM), and k-Nearest Neighbours (KNN) in terms of accuracy and Mean Absolute Error (MAE).

## 1. INTRODUCTION

Facial expression recognition plays a vital role in the artificial intelligence era. According to the human's emotion information, machines can provide personalized services. Many applications, such as virtual reality, personalized recommendations, customer satisfaction, and so on, depend on an efficient and reliable way to recognize the facial expressions. This topic has attracted many researchers for years, but it is still a challenging topic since expression features vary greatly with the head poses, environments, and variations in the different persons involved.

Many methods have been proposed for automated facial expression recognition. Most of the traditional approaches mainly consider still images independently while ignore the temporal relations of the consecutive frames in a sequence which are essential for recognizing subtle changes in the appearance of facial images especially in transiting frames between emotions. Nevertheless, accuracy of the classification or facial expression recognitions is further to be improved. Contributions of this approach are described as follows:

- First, a novel pre-processing method is proposed for the general facial image processing, and it does improve the performance.
- Second, the micro-facial feature extraction methods based on deep learning models with large number of training data is proposed, which can fine-tune the recognition accuracy related parameters in the specific applications.
- Finally, micro-facial expression based deep learning models have the ability to extract more discriminative features which yield in a better interpretation of the texture of human face in visual data. For facial expression recognition, MFEDRL is proposed in this paper. To improve the performance of classifier, this proposed method includes two loss functions such as cross entropy loss function and centre loss function.
- The performance of this proposed approach is evaluated in terms of accuracy and Mean Absolute Error (MAE).

Rest of this paper is organized as follows. Section 2 surveys the previous literature which focused on the research of facial expression recognition. Section 3 describes the material and methods which are used in this paper. Section 4 proposes Micro-Facial Expression Based Deep-Rooted Learning (MFEDRL) classifier. Performance evaluation metrics are described in Section 5 Results of this proposed method are discussed in Section 6. Finally, this paper is concluded with Section 7.

## 2. LITERATURE SURVEY

The conventional facial expression recognition problem becomes even more difficult when we recognize expressions in videos [1]. Most of the times, the entire event of facial expression from the onset to the offset is very quick, which makes the process of expression recognition very challenging [2]. Vo and Le [3] take the second-to-last output layer as the encoded features, and utilize Support Vector Machine (SVM) to be the label predictor. Hamester *et al*. [4] propose a two-channel CNN, and the first convolutional layer in one of the channels is trained by Convolutional Auto-Encoder (CAE) to learn the better capability in order to extract better features. In order to mitigate the effect of head pose, a CNN learns the pose robust features by regressing the features extracted from the Principal Component Analysis Network (PCANet) which has been trained by the frontal facial images with various expressions [5]. Different from traditional learning algorithms in CNN, the model in [6] learns the correlations among the training data. To mitigate the person-specific differences, Meng *et al*. [7] and Zhang *et al*. [8] propose a way to train an identity-aware structure to extract the person-specific features for recognizing the facial expressions. Rather than trying to recognize a single image, Zhang *et al*. [9], Jung *et. al*. [10], and Byeon and Kwak [11] predict the expressions by passing a video, which seems more reasonable in practice; nevertheless, labelling video data is more labour intensive.

Previously researchers extracted facial area based on skin colour as the parameter [12–14] which proved inefficient for gray scale images, then the Haar-like features [12] and AdaBoost algorithm [13,14] used as a cascade of simple features to differentiate face region from nonface region [15]. To eliminate illumination, effect efficient pre-processing technique used in [16]. Gabor wavelet-based feature extraction [17] is employed in effective face recognition but size of feature vector was obtained being large delays effective computation.

Local Binary Patterns (LBP) [18] has proved efficient enough in texture analysis and has been extended to facial feature representation for face recognition [19]. The LBP is easy to use and less in computational complexity. Its robustness to illumination variations proves the significance for face representation inform of histogram. SVM though being a binary classifier was extended to multiclass classification but its complex training system makes it unsuitable for high speed applications. Minimum distance classifiers like Euclidean distance are usually implemented for comparison of the histograms.

Head-pose estimation has many applications, such as social event analysis, human-robot and human-computer interaction, driving assistance, and so forth [20]. Head-pose estimation is challenging because it must cope with changing illumination conditions, variability in face orientation and appearance, partial occlusions of facial landmarks, as well as bounding-box to- face alignment errors. This regression method [21] learns to map high-dimensional feature vectors (extracted from bounding boxes of faces) onto the joint space of head-pose angles and bounding-box shifts, such that they are robustly predicted in the presence of unobservable phenomena [22–24]. The description of the mapping method that combines the merits of unsupervised multiple learning techniques and combinations of regressions.

## 3. MATERIAL AND METHODS

The human brain perceives new concepts with the help of associative links and finding the most similar concept, instead of predefined categorical indexes. That is, a new concept is associated to the most similar concept perceived previously and not by using a naming strategy. Based on this idea, we propose a new method to understand the emotions represented by facial expressions. Our proposed method is in contrast with the traditional approaches that are based on building a complex learning model. In the proposed method, we train separate classifiers for each perceived emotion pattern and collect them. Once a new sample is achieved, each model’s score is calculated, and finally, the one with the highest score is chosen to be the emotion related to the expression. Human Expression approximation process will be developed using four significant steps like, i) Pre-processing using Adaptive Homomorphic Filtering, ii) Micro-feature extraction from face, iii) Proposed MFEDRL models for Deep knowledge system implementation and testing of optimal algorithm were defined, and vi) facial expression estimation based on the final score. At first, the input image is pre-processed to fit for the feature extraction step (Figure 1).

Here, human facial expression estimation methods on facial images are taken into account to a verification system for the proposed work for the system. In pre-processing, such as face detection and rotation rectification, are needed for a given facial image. Rotation rectification can be implemented with the aid of landmarks such as the eyes. Facial expression features are extracted from facial regions appearance variations of a testing image-spatial analysis. The features of motion information are used for expression recognition in a sequence of facial images. Finally, an effective MFEDRL classifier is used to recognize different facial expressions by learning parameters on the extracted features. Then the performance of the algorithm was evaluated using Recognition rate and false measures. The various steps used in the proposed methodology are explained below.

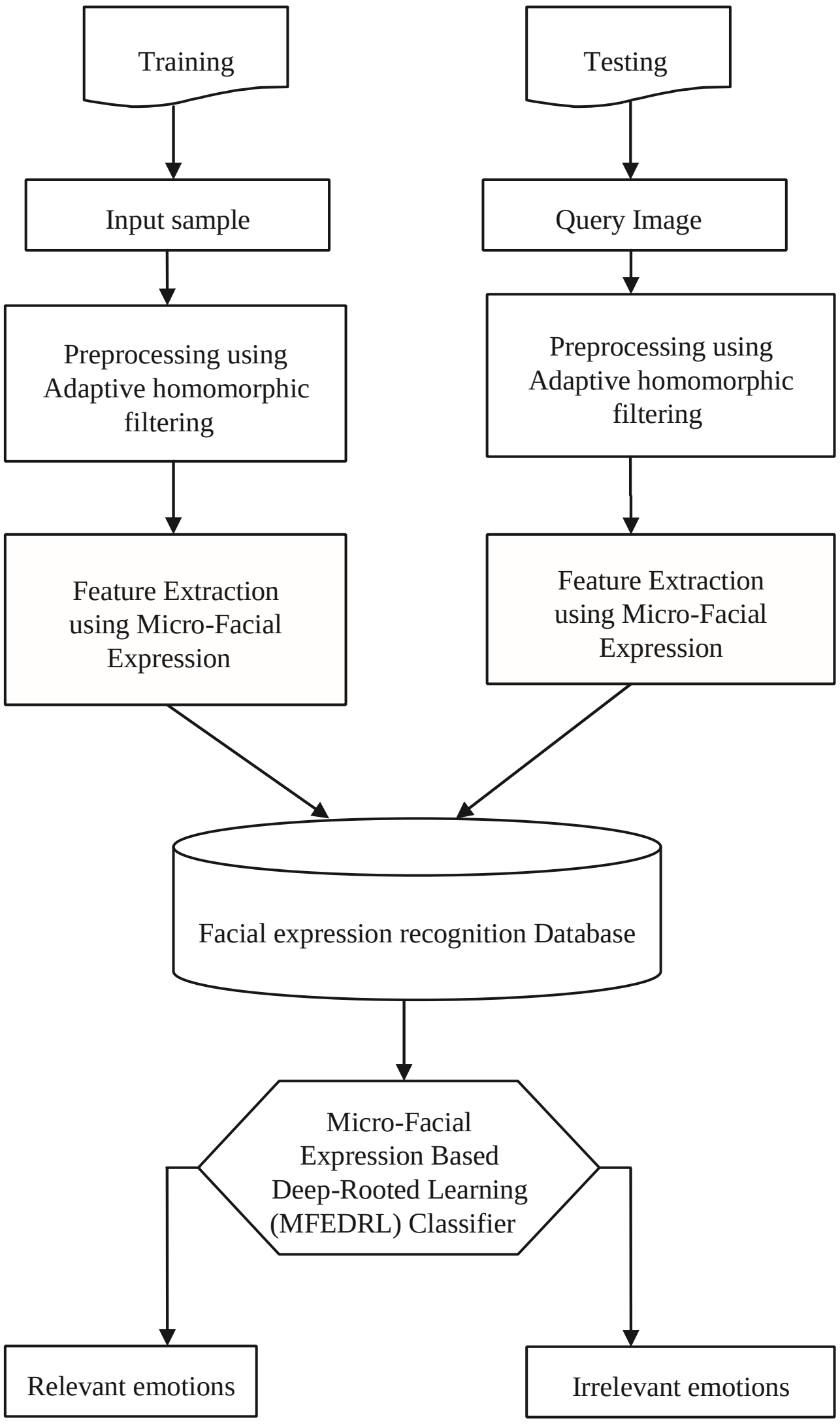

Figure 1: Block Diagram of MFEDRL

### 3.1. Pre-processing Using Adaptive Homomorphic Filter

Pre-processing techniques mainly aimed for the enhancement of the face image without altering the information contained in an image. In this work, an Adaptive Homomorphic Filtering is used for pre-processing processes. Adaptive Homomorphic Filtering [24] is an image enhancement technique that is used to stretch the contrast of the image while attenuating the influence of the illumination on the image content. It is widely used in applications to recognize a face under variant illumination. Based on the understanding of the reflectance model, it is clear that the effect of illumination needs to be reduced and the reflectance part needs to be emphasized in order to normalize the illumination. In other words, attenuate the low frequency components and strengthen the high frequency components. To do so, Homomorphic Filtering (HF) is used.

In Adaptive Homomorphic Filtering [26], the output of the image is weighted with a sum of input pixels. In general, the modified Gaussian high pass filter can be used to define $H\ (u,\ v)$ as follows:

$$Adaptive\ Homomorphic\ Filtering = H\left(x, y\right) = (\gamma H - \gamma L)\left[1 - \exp^{-c\left(\frac{D^2(x,y)}{D_0^2}\right)}\right] + \gamma L \quad (1)$$

where $\gamma H > 1$ and $\gamma L < 1$ are the parameters of the filter, $D_0$ is the cut-off frequency, C is a constant that controls the increasing degree of exponential function, and $D\ (x,\ y)$ is the distance from $(x,)$ to the origin of the cantered Fourier transform. 3.1.1. Adaptive homomorphic filter algorithm steps

**Procedure:**

Input: Null

Output: Preprocessed image Pimg.

Start Capture Image Img.

Pre-process the image Img.

Identify a set of all unique intensity values of image Timg. Iv = ∑Intensity Values (Timg) ₵Iv

For each intensity value $Iv_i$ from Iv Compute probability distribution

$$pdf = \frac{\sum Pi\ (Timg) == Ivi}{Size\ (Iv)}$$

Restore the image pixel according to pdf. End. Stop

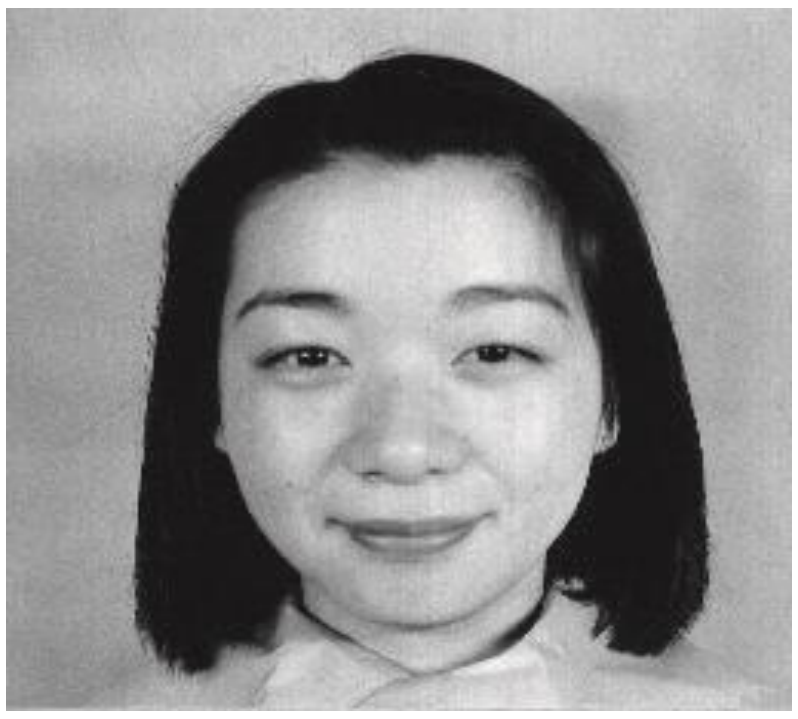

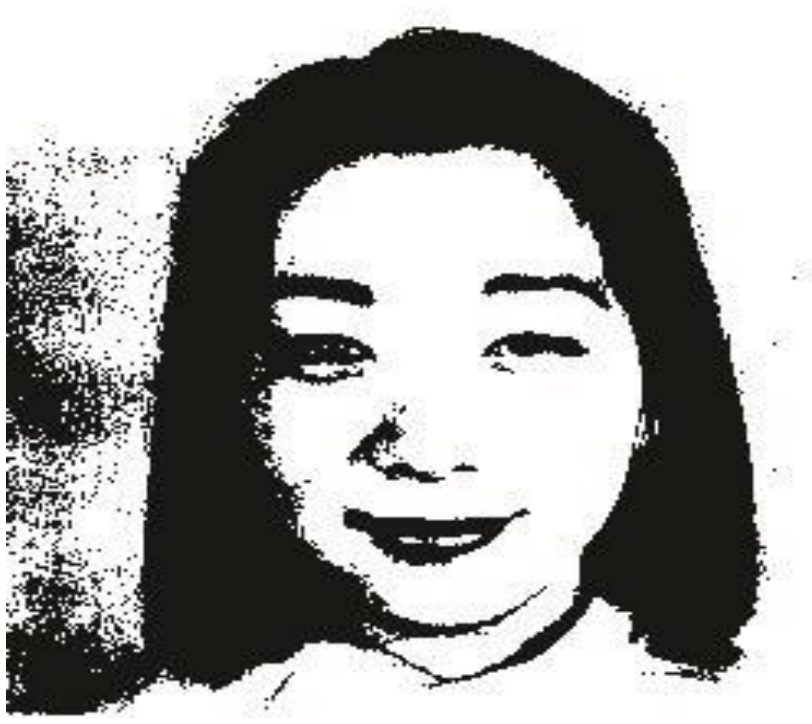

2a) Input image 2b) Pre-processed image

**Figure 2** Result of pre-processing.

The above Figure 2 shows result of an input image and its pre-processed image using Adaptive Homomorphic Filtering. As compared to input image the pre-processed image shows illumination noise based image. The above-discussed algorithm performs the histogram equalization of the input image captured through the device and improves the quality of input image obtained.

## 3.2. Micro-Facial Feature Extraction

After preprocessing, the micro-facial features are extracted from the eye regions (Reye) facial region (RFace), and the mouth region (Rmouth). The first region for the eyes region (Reye): To extract facial local features from eyes regions, we use a small template in the channel Reye. Reye first crops the above about one-third of facial regions as the eyes' region, then averagely pools the eyes regions to size 140 × 40.

The second channel Facial region (RFace) for the whole face: RFace is used to extract and learn the global micro-deep facial feature from the whole facial region. It first averagely normalizes the input data *RFace* (facial region) to the size of 200 × 200. The third channel RFace for the mouth region: Rmouth extracts the local mouth convolutional features. It first crops the below about one-third of facial region as the mouth region, and resizes the mouth region as 140 × 40.

906

### 3.2.1. Algorithm

**Input**: Convolutional feature matrix sRj, with *j*- {Reye, RFace, Rmouth}, Modified weight matrix s$w_i^j$,

Reset image vectors $Image_i^j$.

**Output**: Micro-facial feature (MFF) vector

**Process**

Step 1: Compress eye, face and nose & Mouth matrix s in each channel:

$y_i^j$ = Reshape Rj->{Reye, RFace, Rmouth},

SET = minimum diameter of $C_f$;

Step 2: Calculate $\left(y_i^j \cdot w_i^j +\right.$ and stimulate the linked level parameters (LP):

$f_i^j$ = Rectified Linear Unit$Image_i^j$)

Step 3: the combination of micro-deep high-level features and activation function $a^1$ = Contat ($f^1$, $f^2$)

$p^1$ $\left(a^1.w^1 +\right.$ = Rectified Linear Unit $Image^1$)

Step 4: $(p^1, f_i^3)$ Estimate the feature vectors

Fuse and enhance high-level features in the second fusion layer:

$a^2$ = Contat

MFF = Rectified Linear Unit ($a^2 \cdot w^2 + Image^2$)

Step 5: RETURN MFF.

Step 6: Construct micro-facial feature Vector values MFFv = $\int_{j=1}^{k} (V1 \cap V2 \cap V3) \in Ds\,(j)$

Step 6: Stop.

(a) **LBP features**

The LBP feature extraction [27] method is a simple and nonparametric method that describes spatial information of the pixels regarding to their neighbour pixels. This is done by assigning a label (decimal value) to each pixel using Equations (2) and (3). The labels are then used to form a histogram in order to represent the image.

$$LBP = \sum_{i=0}^{n-1} s\left(g_i - g_c\right) 2^i \qquad (2)$$

And,

$$s(x) = \begin{cases} 1 & if\ x \geq 0 \\ 0 & if\ x < 0 \end{cases} \qquad (3)$$

In Equation (2), $g_i$ and $g_c$ indicate gray-level of the neighbourhood pixels and the centre pixel, respectively, and n is the number of neighbours.

(b) **Histogram of Oriented Gradients (HOG) features**

The HOG feature extraction [28] method uses local gradients to describe the shape of an object. To this end, the horizontal and vertical gradients of a given image are calculated first. Then, the magnitude ($q$) and the direction ($\theta$) of each pixel's gradient are found as follow:

$$q = \sqrt{g^2{}_x + g^2{}_y} \qquad (4)$$

$$\theta = \arctan\left( g_y / g_x \right) \qquad (5)$$

Where, $g_x$ and $g_y$ indicate horizontal and vertical gradients, respectively.

Then, in the step of micro-deep facial feature extraction, MFF integrates the deep convolutional features into a high-level representation. The discriminative illustration is shared by the subnetworks in the following Micro-Facial Expression Recognition Based Deep-Rooted Learning (MFEDRL) Facial expression recognition network. Finally, to suppress the errors from MFEDRL consists of multiple specific recognition branches and an additional pose estimation subnetwork. MFEDRL is optimized by minimizing a loss of LBP and HOG features based conditioned expression recognition loss. Then these features are moved to classification.

## 4. MICRO-FACIAL EXPRESSION RECOGNITION BASED DEEP-ROOTED LEARNING CLASSIFICATION

MFEDRL facial expression recognition and classification is a supervised strategy which depends on probability spreading models that could be parametric or nonparametric classifier models. In spite of the fact that, in proposed classification is provided with a gathering of marked (pregrouped) images and the issue is to name recently experienced, unlabelled images and the algorithm steps are examined below. Figure 3 shows the block diagram of the proposed MFEDRL classification. Neural networks trained on a limited set of emotions generally tend to result in higher rates of accurate classifications. To give the most extreme example, training on a dataset containing only examples of happiness and anger tends to produce very high accuracy.
MFEDRL Training is simultaneously learning multiple classification/prediction tasks that are related to one another. In other words, in the framework of MFEDRL Training, multiple related classification tasks are learned jointly and information is shared across

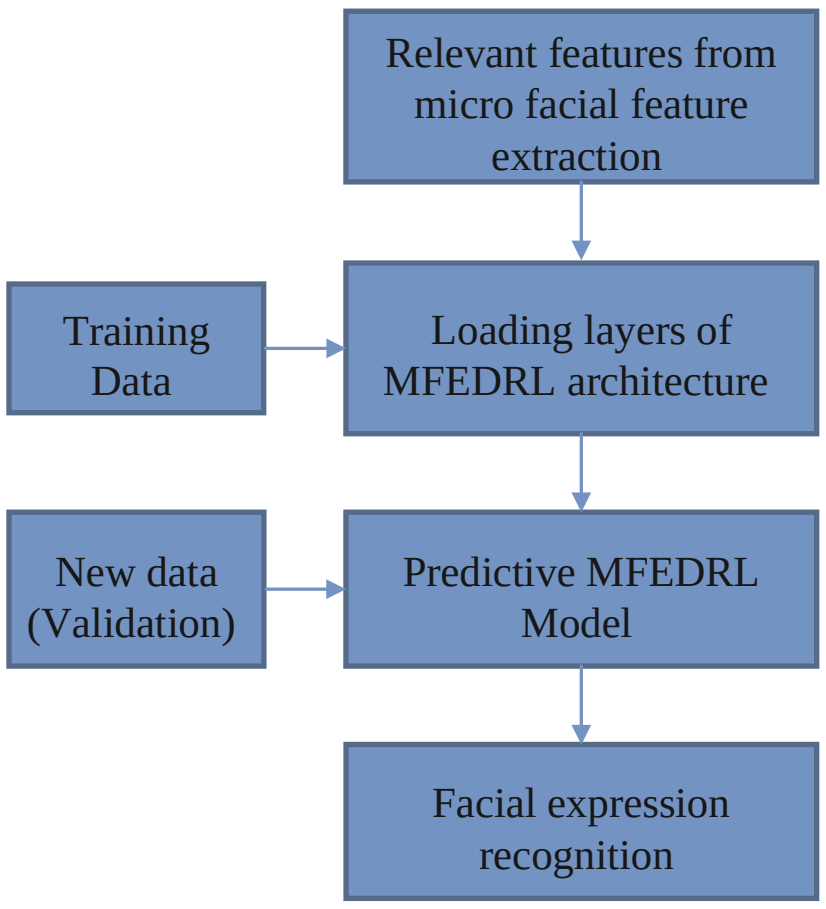

**Figure 3** Block diagram of the proposed Micro-Facial Expression Based Deep-Rooted Learning (MFEDRL) classification.

the tasks. In the first part, knowledge is shared or learned jointly among many tasks. In the second part, information or knowledge for the specific task is learned separately. In other words, common knowledge learned in the first part is adapted to a specific task in the second part. When generalized knowledge is further adapted to the specific task, modelling process results in better optimization effect. To take advantage on MFEDRL classifier modelling algorithm steps were discussed as follows:

**Algorithm: MFEDRL Training**

Start

Initialize all weights and biases of the MFEDRL to a small value. Set learning rate such that <1, n = 1

repeat

for m = 1 to M do

propagate pattern $x_m$ through the network for k = 1 to the number of neurons in the output layer

Find error end for for layers L-1 to 1 do for maps j = 1 to J do find error factor to be back propagated end for end for for i = 1 to L do for j = 1 to J do for all weights of map, j do Update weights and biases end for end for end for

n = n + 1

Find Mean Square Error (MSE), Until MSE1< or n > maximum
bounds, Then Transfer Knowledge Phase repeat
for tk = 1 to TK (number of new training samples) propagate pattern $x_{tk}$ through the MFEDRL network for z = 1 to the number of neurons in the last MFEDRL layer(Z) find output Oz of last layer of the MFEDRL layer. Oz = (O1, O2, O3 ........ OZ) Find output using MFEDRL framework
end for
Weight Update learning phase for transfer learning phase using
MFEDRL n = 1
repeat
for tk = 1 to TK Train the MFEDRL layers using end for
n = n + 1
Find MSE
Until MSE< or n> maximum bounds
End

### 4.1. Micro-Facial Expression Based Deep-Rooted Learning (MFEDRL Algorithm Steps

Due to large variations across subjects, facial expression recognition performance degrades on unseen subjects in real-world scenarios. To cope with this problem, proposed MFEDRL is used to construct a subspace with several expressions of the same subject, and then a MFEDRL based Convolutional Neural Networks (CNNs) is used for facial expression recognition. The last fully connected layer of Facial Expression Recognition (FER)-net is used as features for both the input image and the generated images, and the input image is labelled as one of the six basic facial expressions based on a distance function in the feature space:

$$Pr\,edict = \arg\min_{i} \lVert Feat\,(I_{input} - Feat\,(I_i))\rVert \quad (6)$$

Here, *Feat* (·) is a feature extraction function. Any method, except the FER-net used here, that can be used to extract the facial expression related features works as well.

The proposed Deep Learning Network is used to find out the facial Expression estimation, and it is prepared by utilizing the above primitive highlights esteems which are coerced from each and each picture. The proposed Deep Learning Network is all around ready to use the forced highlights. The original MFEDRL is a system to three information units, n concealed units and one yield unit. The contribution of the neural system is the element vectors are separated from the pictures. The system is prepared with a substantial arrangement of various highlights from input confront images to empower them to gauge the Expression factor in the testing stage viable. The proposed Deep Learning Network works making utilization of two steps, one is the preparation stage, and the other is the awkward stage.

#### 4.1.1. MFEDRL algorithm steps

The accompanying algorithm steps are taken after entirely in this work.

**Step-1:** The pre-processing step is quite simple: firstly, normalizing data per image, and then normalizing data per pixel. And Pre-processing step is applied to both training data and test data.

**Step-2:** For each training image, we then subtract from each pixel its mean value, and then set the standard deviation of each pixel over all training images to 1.

**Step-3:** Due to the small amount of training data, we apply learning techniques to increase the amount of training samples in order to avoid over fitting and improve recognition accuracy. For each image, we perform the following successive transforms:

i. Mirror the image with a probability of 0.5.

ii. Rotate the image with a random angle from −45 to 45 (in degrees).

iii. Rescale the image, whose original size in the dataset is 48 × 48, to a random size in the range from 42 × 42 to 54 × 54. iv. Take a random crop of size 42 × 42 from the rescaled image.

**Step-4:** In the training phase, we minimize the loss function by using mini-batch gradient descent with momentum and the backpropagation algorithm. The batch size is 256, the momentum is 0.9.

**Step-5:** To avoid over fitting, we apply the dropout technique to the fully connected layers (except for the output one) with a dropout probability of 0.5. During training phase, we use a strategy that decreasesthelearningrate10timesifthetraininglossstopsimproving. The experiments show that the learning rate is often decreased about five times, and the training phase is often finished after about 1,400 epochs.
**Step-6:** All biases are initialized by zero. All weights are randomly initialized from a Gaussian distribution with zero mean and the following standard deviation:

$$\delta = \sqrt{\frac{2}{nInput}} \qquad (7)$$

Where Convolutional layer, nInput = filter size × filter size × the depth of the previous layer. For the fully connected layer, nInput = the number of neurons in the previous layer. For each iteration, the next 256 images are taken from the training data. After performing training, we will have a batch of 256 augmented images, each of which has size of 42 × 42. These images are then fed to the network for training. After each epoch, the training data is randomly shuffled.

**Step-7:** As mentioned above, we try to use different loss functions [29]: cross-entropy and center loss function. The cross-entropy loss function is defined as follows:

$$H = -\frac{1}{N}\sum_{i=1}^{N}\sum_{j=1}^{C} y_i(j)\log\left(\hat{y}_i(j)\right) \qquad (8)$$

where: $y_i(j)$ indicates whether $j$ is the correct label of $i$th image, and $N$ is the number of images in the training data; $C$ is the number of emotion classes, $y_i(j)$ expresses the different emotion based probability that $j$ is the correct label of $i^{th}$ image.

Also, centre loss function is introduced in this proposed method for improving the discriminative energy of the deeply learned features. This loss function is defined as follows:

$$L_C = \frac{1}{2}\sum_{i=1}^{n} ||x_i - C_{y_i}||^2 \qquad (9)$$

–

Where, $x_i$ represents the $i$th deep feature in the class of $y_i$, $C_{yi}$ represents the $y_i$th class center of deep features.

**Step-8:** In the test phase, having a trained network and an input image of size 48 × 48, we use multiple crops to predict the true class label.

The cross-entropy loss works fine when the predicted labels are mutually exclusive. However, the labels of the dataset are interrelated. To address this problem, participants adopted the center loss function as an auxiliary loss to reduce the effect of similar label. The center loss can simultaneously learns each class center of deep features and penalizes the distances between the deep features and their corresponding class center. This loss enhances the ability of model to distinguish similar samples and improves the overall performance.

Figure 4 shows the overall work process of the proposed system. The performance assessment of the improvement techniques that are utilized to build up the last distinguishing evidence framework is a fundamental advance in any examination. Distinctive scientists utilize different parameters for investigation. This segment introduces the different measurements used to assess the techniques proposed in each period of the examination. The execution of the classification has assessed with respect to classification exactness, accuracy, review and F-measure from the perplexity network of classification.

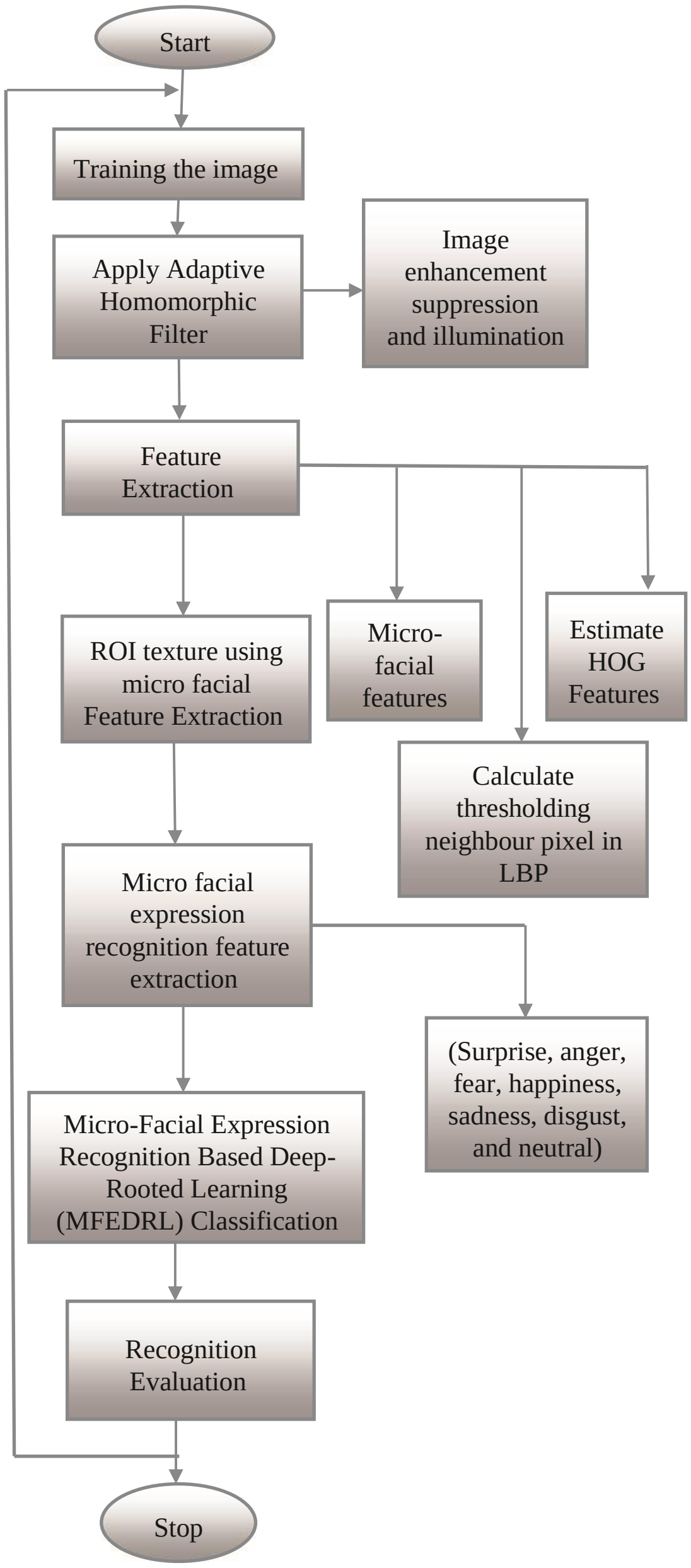

Figure 4: Flow chart for Micro-Facial Expression Recognition Based Deep-Rooted Learning (MFEDRL) classifier based facial expression recognition.

In the training phase, the input image is feature extracted and this feature vector is given as the input to the MFEDRL network. Initially, the nodes are given random weights. As the output is already known in the training phase, the output obtained from the MFEDRL network is compared to the original and weights are varied so as to reduce the error. This process is carried for large data so as to yield a stable system having weights assigned in the nodes.

## 5. PERFORMANCE ANALYSIS

The performance of a proposed method of the face and feature points Expression estimation is calculated by paying the numerical events mentioned below.

### 5.1. Mean Absolute Error

A normal way of computing the accuracy of a regressor is to regular the complete deviation of each example's sticker from its estimated value. More formally, MAE of a given data set is calculated as follows:

$$MAE = \frac{1}{N}\sum_{i-1}^{N}(x_i - x_i) \qquad (10)$$

Where, $x_i$ is the right tag, that is, the regular of apparent Expression annotations for sample $i$, $x_i$ is the predicted value, and $N$ is the number of challenging samples.

### 5.2. Precision

Precision is the fraction of signals recognized that are relevant to the query input.

$$precision = \frac{TP}{TP + FP} \qquad (11)$$

### 5.3. Recall

Recall determines the portion of signals that are of interest to the query input signals that are successfully recognized.

$$recall = \frac{TP}{TP + FN} \qquad (12)$$

### 5.4. F-Measure

F-measure is a weighted harmonic mean of precision and recall.

$$FMeasure = 2\frac{precision \times recall}{precision + recall} \qquad (13)$$

### 5.5. Accuracy

Accuracy measures how close the recognized image is to the query input.

$$Accuracy = \frac{(TP + TN)}{(TP + FP + TN + FN)} \qquad (14)$$

### 5.6. Sensitivity

Sensitivity events the amount of information that is relevant to the query input that is successfully recognized.

$$Sensitivity = \frac{TP}{TP + FN} \quad (15)$$

### 5.7. Specificity

Specificity events the fraction of feedback signals that are related to the query input that is not correctly identified.

$$Specificity = \frac{TN}{FP + TN} \quad (16)$$

Facial expression recognition process is tested with different datasets of input face and feature points, and the future result of the proposed system has been shown below.

## 6. RESULTS AND DISCUSSIONS

The implementation of the proposed micro-facial expression recognition approach is done on the working platform of MATLAB (version 2017a). The input for the micro-facial expression recognition is chosen from open source databases and these are handled by our proposed system. The information flag is sifted utilizing for pre-processing. For each pre-processed flag separated, extended highlights are extricated. Once the element extraction is done, the last stage is the Expression estimation of face inquiry. The system utilized here for the micro-facial expression recognition using streamlining strategy called MFEDRL calculation. Finally, the micro-facial expression recognition focuses are perceived from the info using the proposed procedure. The experimental result is shown in the below sections. Table 1 lists the details of evaluation parameters utilized for the verification of the methods proposed. For microfacial expression recognition, The Japanese Female Facial Expression (JAFFE) Database is utilized. This database consists of 213 images of seven facial expressions of 10 Japanese female models.

The micro-facial expression recognition image from face image has been shown in Figure 5. The testing dataset takes the single object as input. The input image is pre-processed then micro-facial expression features were extracted and trained by using MFEDRL. Then Performance factors are estimated for the comparisons. The performance of the existing methods such as *k*NN (*k*-Nearest Neighbours), SVM, ANN (Artificial Neural Network), DNN (Deep Neural Network), and CNN (Convolutional Neural Network) are compared with that of the proposed micro-facial expression recognition approach of MFEDRL.

As shown in the Table 2, same values are attained for recall, F-measure, precision, and sensitivity of the proposed MFEDRL. Also, accuracy of the proposed MFEDRL classifier is improved than that of the existing classifiers as data is given in the table. Because of the loss functions such as cross entropy and centre loss functions, the proposed MFEDRL attains the accuracy value to 98%. Figure 6a shows the graphical representation of the comparison of sensitivity, specificity, and accuracy of different classifiers. Similarly, Figure 6b shows the graphical representation of the comparison of precision, recall, and F-measure of different classifiers.

| **Expressions** | **Women 1** | **Women 2** |
|---|---|---|
| Fear | 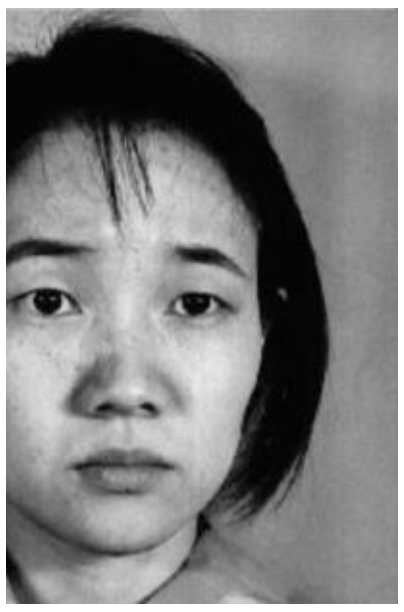 | 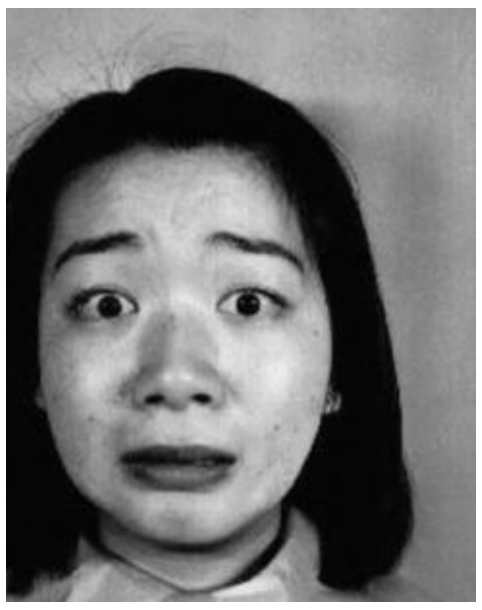 |
| Happy | 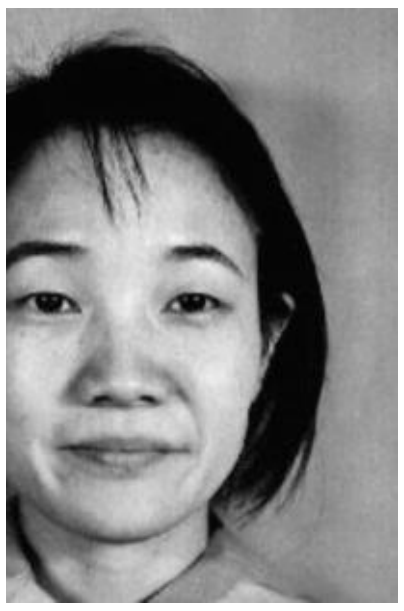 | 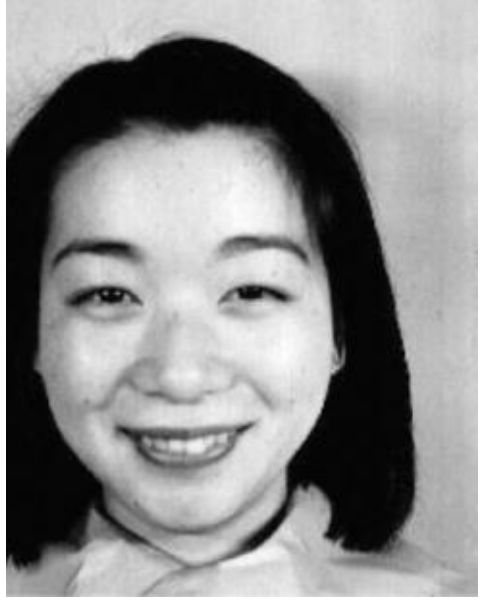 |
| Sad | 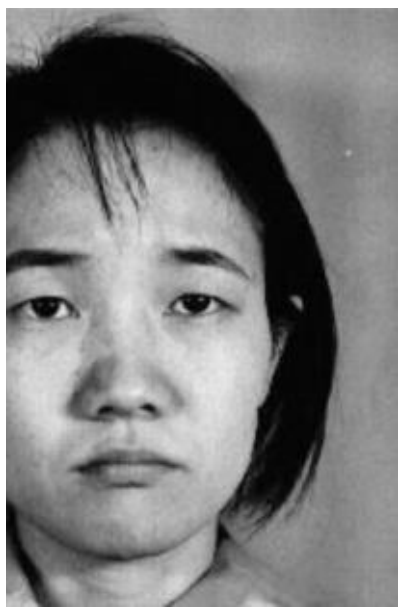 | 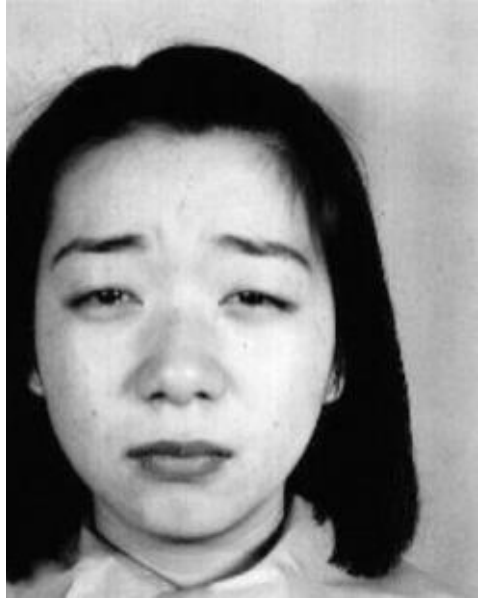 |
| Surprise | 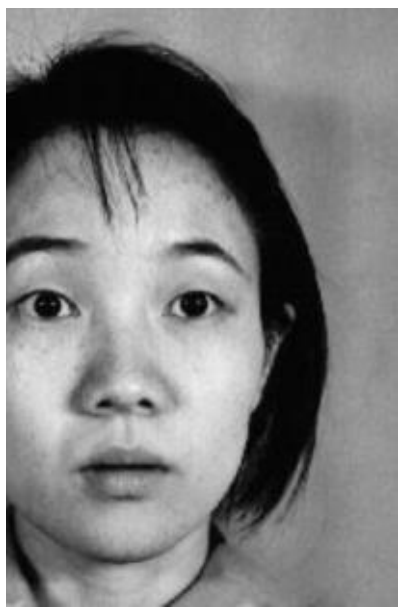 | 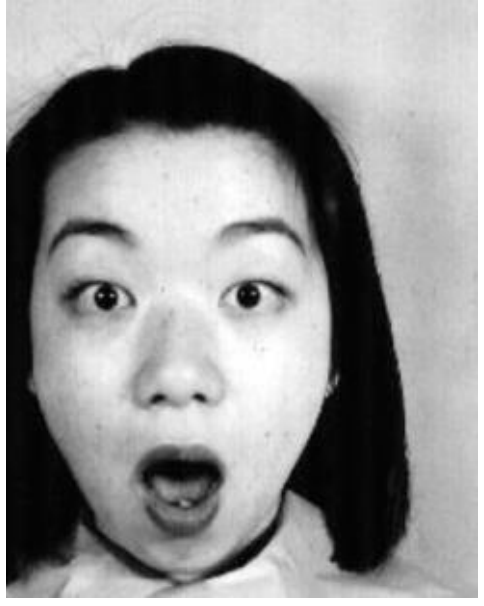 |
| Angry | 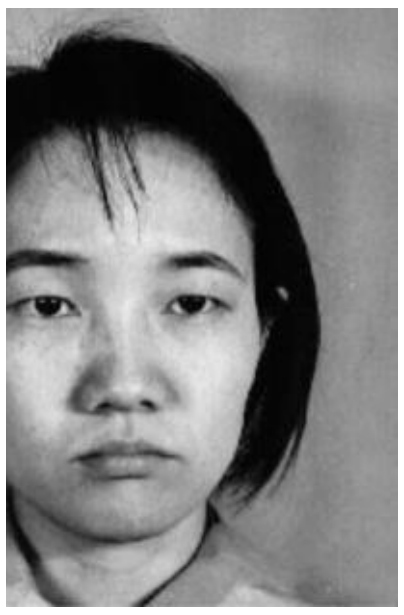 | 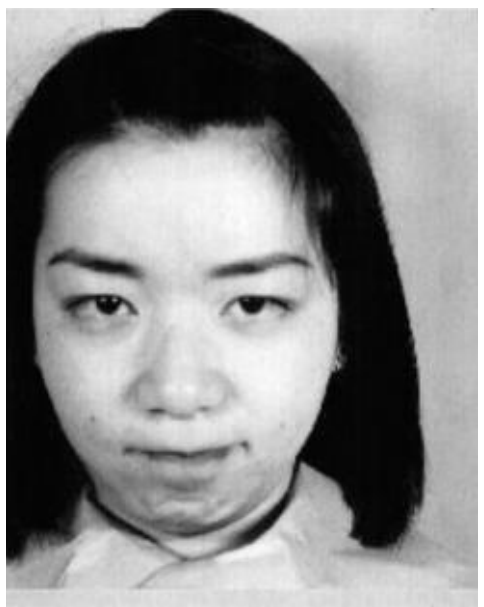 |

**Figure 5** Proposed facial expression recognition classifier based output. (*Continued*)

| **Expressions** | **Women 1** | **Women 2** |
|---|---|---|
| Digust | 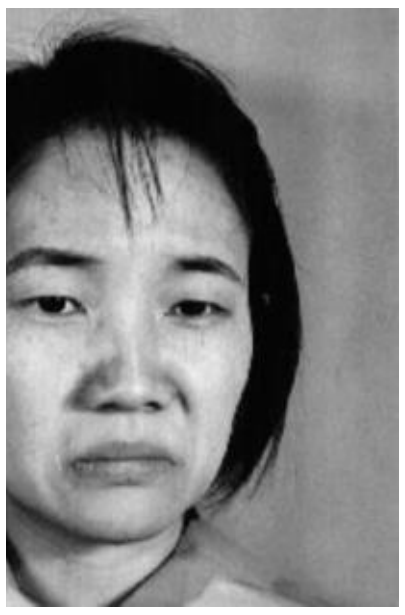 | 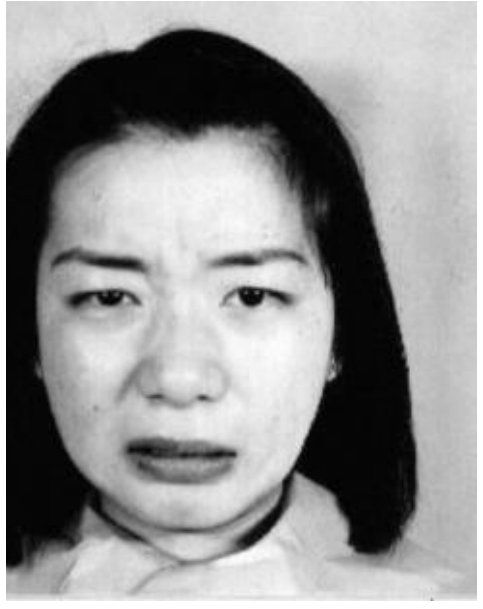 |
| Neutral | 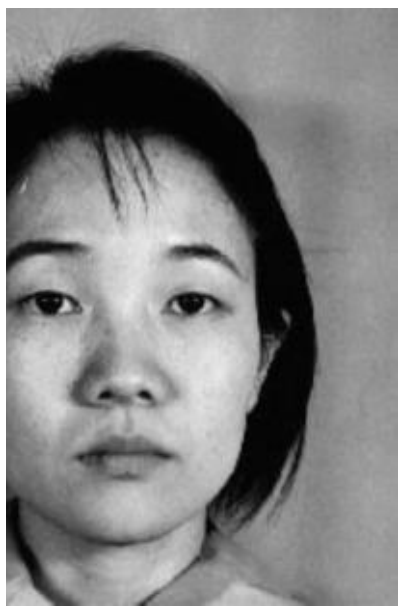 | 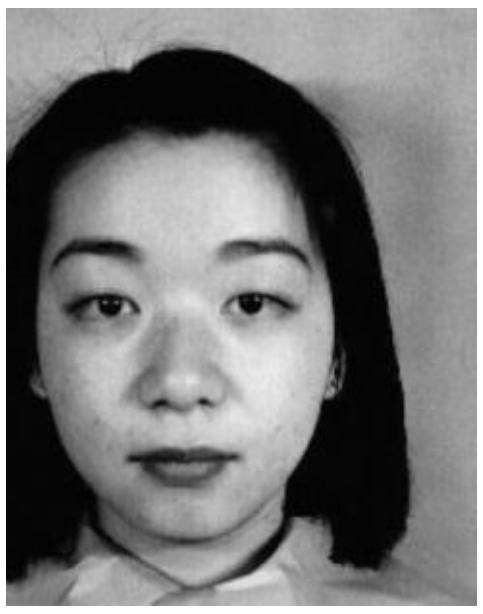 |

**Figure 5** Proposed facial expression recognition classifier-based output.

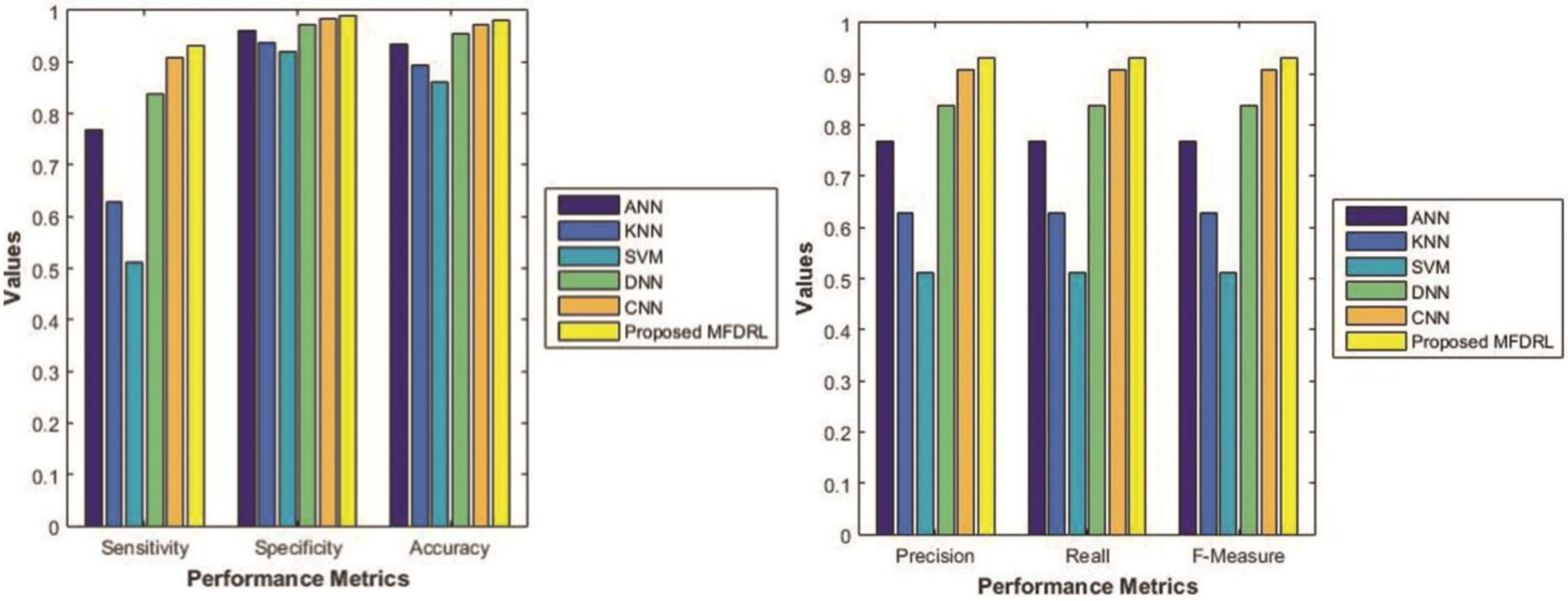

(a) (b)

**Figure 6** (a & b) Performance comparison of various methods.

**Table 1** Details of evaluation parameters.

| Parameter | Value |
|---|---|
| Data set | JAFFE |
| No. of persons | 10 women |
| No. of face images | 213 facial expression images |
| Tool used | MATLAB 2017a |

JAFFE, Japanese Female Facial Expression.

**Table 2** Classifier performance comparison for various methods.

| Measures(%) | MFE DRL | CNN | DNN | ANN | KNN | SVM |
|---|---|---|---|---|---|---|
| Accuracy | 0.98 | 0.97 | 0.95 | 0.93 | 0.89 | 0.86 |
| Recall | 0.93 | 0.91 | 0.84 | 0.77 | 0.63 | 0.51 |
| F-measure | 0.93 | 0.90 | 0.83 | 0.76 | 0.62 | 0.50 |
| Precision | 0.93 | 0.90 | 0.83 | 0.76 | 0.62 | 0.50 |
| Sensitivity | 0.93 | 0.91 | 0.84 | 0.77 | 0.63 | 0.51 |
| Specificity | 0.99 | 0.98 | 0.97 | 0.96 | 0.94 | 0.92 |

ANN, Artificial Neural Network; CNN, Convolutional Neural Network; DNN, Deep Neural Network; KNN, *k*-Nearest Neighbours; MFEDRL, Micro-Facial Expression Based Deep-Rooted Learning; SVM, Support Vector Machine.

The comparison graph in Figure 7 shows the proposed micro-facial expression recognition based MFEDRL is higher than the existing method regarding accuracy, sensitivity, specificity, precision, recall, FM. From Figure 7, it is noted that the proposed micro-facial expression recognition MFEDRL method increases a percentage of accuracy with reduced error level when compared with existing methods. Table 3 shows the comparison of MAE of the various methods. As shown in the table, MAE of the proposed method is reduced than that of the existing methods.

**Table 3** MAE for various methods.

| Techniques Used | Mean Absolute Error |
|---|---|
| KNN | 0.90 |
| SVM | 1.42 |
| ANN | 0.32 |
| DNN | 0.46 |
| CNN | 0.18 |
| Proposed MFEDRL | 0.069 |

ANN, Artificial Neural Network; CNN, Convolutional Neural Network; DNN, Deep Neural Network; MAE, Mean Absolute Error; MFEDRL, Micro-Facial Expression Based Deep-Rooted Learning; SVM, Support Vector Machine.

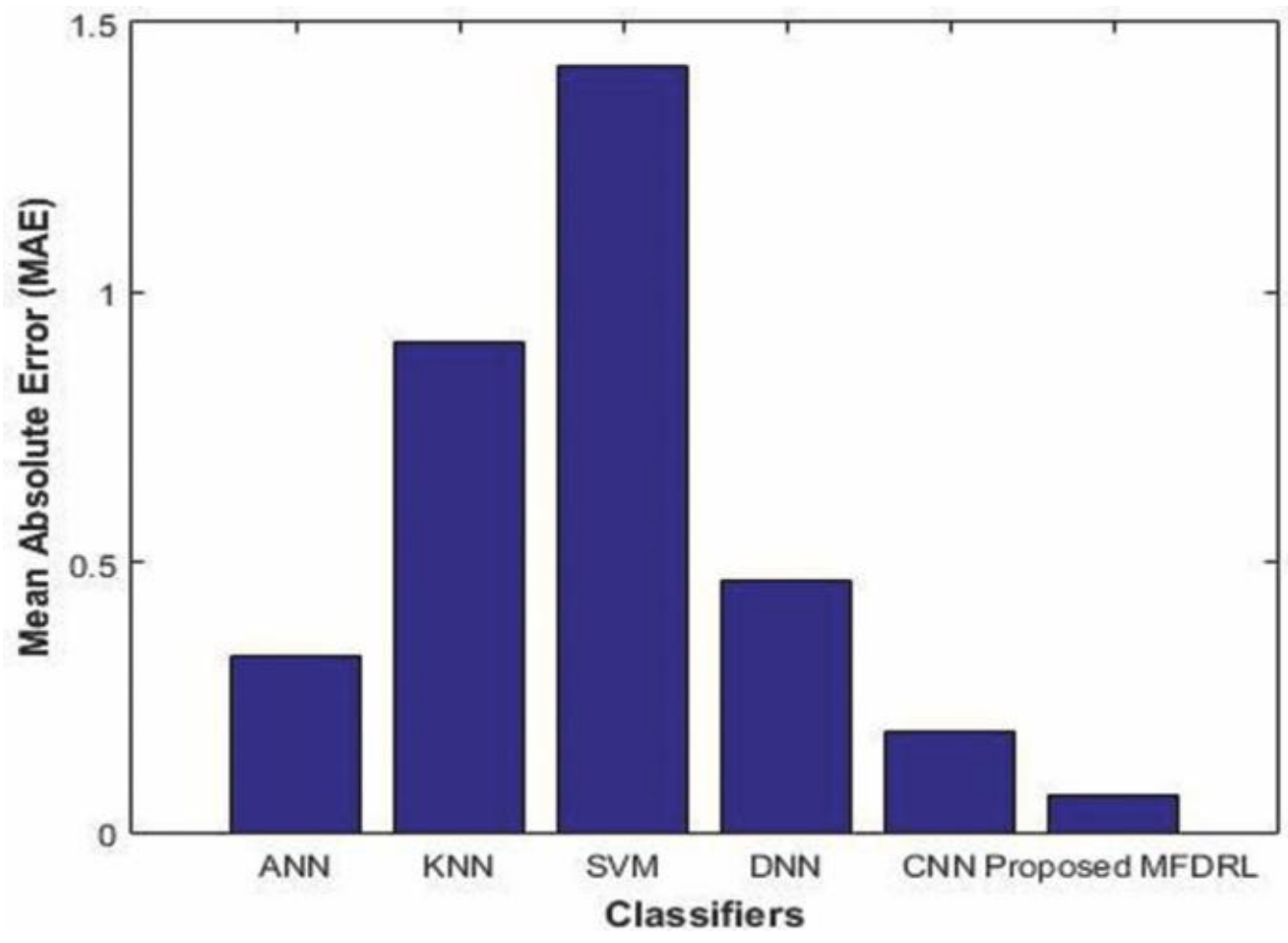

Figure 7: Mean Absolute Error (MAE) comparison of different facial expression estimation methods on the Japanese Female Facial Expression (JAFFE) Dataset.

The comparison graph in Figure 7 shows the MAE comparison of different facial expression estimation methods and the proposed method MFEDRL–MAE error has lower than the existing method.

## 7. CONCLUSION

In this work, the proposed approach is aimed to recognize the micro-facial expression recognition from the learned image dataset. From above results, it is clear that the proposed MFEDRL approach can classify better. Our proposed approach is experimented only with a limited number of images in the database. It can be extended by increasing the number of images in the dataset and images from various databases. The experimental results describe the efficiency of the micro-facial expression recognition by computing the correct and incorrect estimation with the MAE results. Results predict the level of error, and also the micro-facial expression recognition accuracy is obtained in JAFFE databases and depicts that the proposed approach received best score than the existing methods. All the methods have been evaluated for their efficiency and have produced useful results. Further, the research can be improved by adapting various similarity measures in computing the similarity of the micro-facial expression recognition.